\documentclass[journal]{IEEEtran}

\usepackage[utf8]{inputenc} 
\usepackage[T1]{fontenc}    
\usepackage{hyperref}       
\usepackage{url}            
\usepackage{booktabs}       
\usepackage{amsfonts}       
\usepackage{nicefrac}       
\usepackage{microtype}      
\usepackage{graphicx}
\usepackage{color}
\usepackage{multirow}
\usepackage{booktabs}
\usepackage{enumerate}
\usepackage{amsmath}

\ifCLASSINFOpdf
\else
\fi
\begin{document}
%
\title{Fully Convolutional Online Tracking}
%
%
%

\author{Yutao~Cui,
        Cheng~Jiang,
        Limin~Wang,~\textit{Member},~\textit{IEEE},
        Gangshan~Wu,~\textit{Member},~\textit{IEEE} 
\thanks{Yutao Cui, Cheng Jiang, Limin Wang and Gangshan Wu are with the State Key Laboratory for Novel Software Technology, Nanjing University, Nanjing, 210023 China (e-mail:cuiyutao@smail.nju.edu.cn; mg1933027@smail.nju.edu.cn; lmwang@nju.edu.cn; gswu@nju.edu.cn).}
}

\maketitle

\begin{abstract}
Online learning has turned out to be effective for improving tracking performance. However, it could be simply applied for classification branch, but still remains challenging to adapt to regression branch due to its complex design and intrinsic requirement for high-quality online samples. To tackle this issue, we present the fully convolutional online tracking framework, coined as FCOT, and focus on enabling online learning for both classification and regression branches by using a target filter based tracking paradigm.
Our key contribution is to introduce an online regression model generator (RMG) for initializing weights of the target filter with online samples and then optimizing this target filter weights based on the groundtruth samples at the first frame.
Based on the online RGM, we devise a simple anchor-free tracker (FCOT), composed of a feature backbone, an up-sampling decoder, a multi-scale classification branch, and a multi-scale regression branch.
Thanks to the unique design of RMG, our FCOT can not only be more effective in handling target variation along temporal dimension thus generating more precise results, but also overcome the issue of error accumulation during the tracking procedure. 
In addition, due to its simplicity in design, our FCOT could be trained and deployed in a fully convolutional manner with a real-time running speed.
The proposed FCOT achieves the state-of-the-art performance on seven benchmarks, including VOT2018, LaSOT, TrackingNet, GOT-10k, OTB100, UAV123, and NFS. Code and models of our FCOT have been released at: \url{https://github.com/MCG-NJU/FCOT}.
\end{abstract}



%
\IEEEpeerreviewmaketitle

\section{Introduction}
\IEEEPARstart{V}{isual} object tracking~\cite{dimp,siamfc,atom,siamrpn} is a fundamental task in computer vision, which aims at estimating the state of an arbitrary target in every frame of a video, given its bounding box in the first frame. It has a variety of applications such as human-computer~\cite{introduction1} and visual surveillance~\cite{introduction2}.
However, tracking still remains as a challenging task due to several factors such as illumination change, occlusion, and background clutter.
In addition, target appearance variation along temporal dimension will further increase difficulty for robust tracking.

A typical tracker~\cite{atom} comprises a classification branch, to locate the target coarsely by discriminating it from the background, and a regression branch, to generate the accurate bounding box of the target. 
For classification task, the current approaches could be divided into generative trackers (e.g. SiamFC~\cite{siamfc}) and discriminative trackers (e.g., DiMP~\cite{dimp}) from the perspective of whether explicitly performing discriminative learning on online samples. The generative tracker typically employs a fixed target template without modeling background clutter, while the discriminative tracker learns an adaptive filter by maximizing the response gap between target and background. It is well established that this discriminative training would increase the robustness of tracking~\cite{dimp}.
For regression task, the existing methods usually depend on hand-crafted design, such as anchor box placement~\cite{siamrpn,siamrpnPlus,dasiamrpn}, or box sampling and refinement~\cite{dimp}. Due to its complex design, this regression branch cannot be easily optimized with online learning for each tracked target, and thus could not be adaptive for each target to handle object deformation. 
Besides, there is intrinsic demands for high-quality training samples in regression. However, the predictions of online samples may be imprecise due to the error accumulation, which cause limitation to online regression. 

\begin{figure}[t]
\centering
\includegraphics[width=\linewidth]{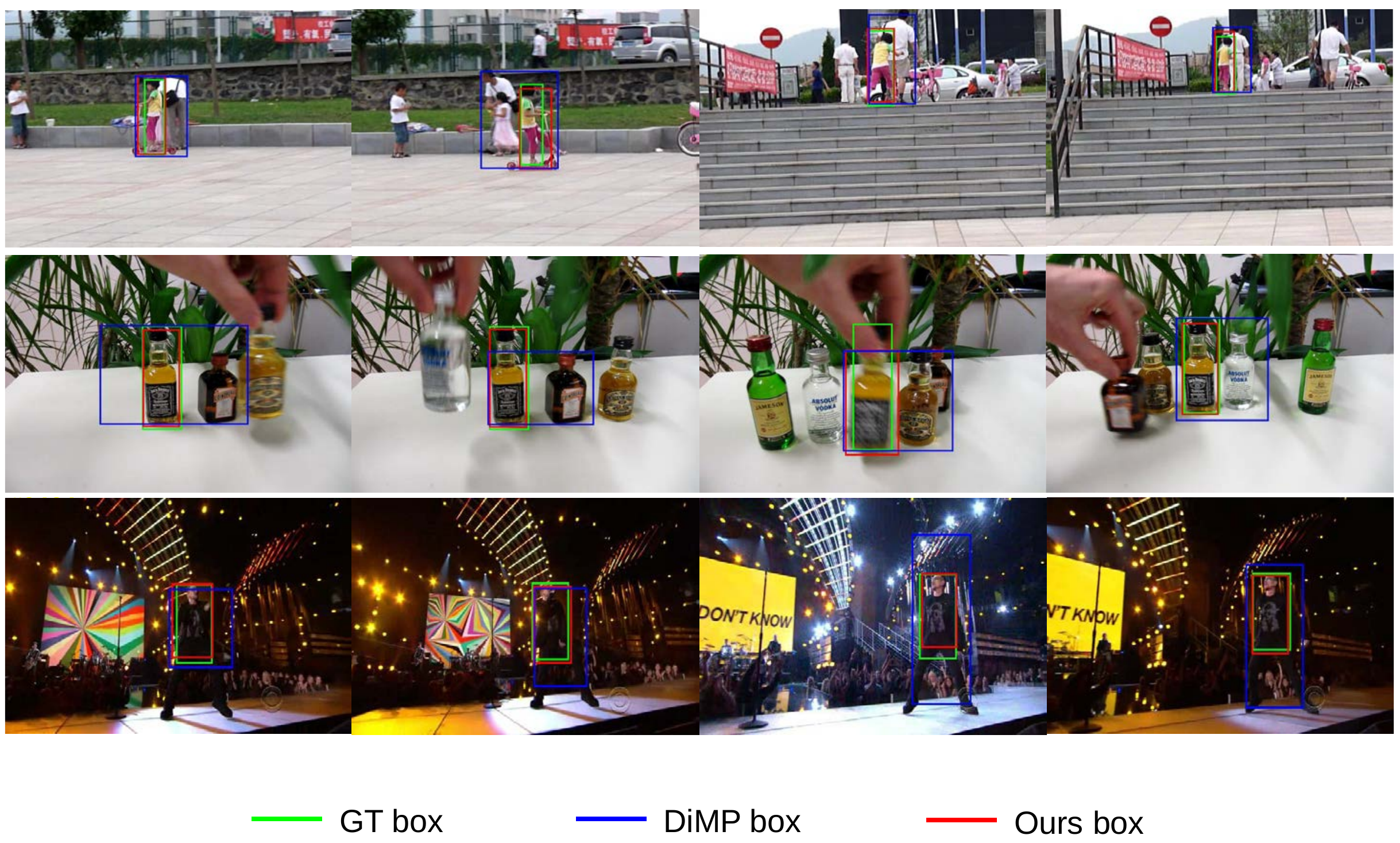}
\caption{A comparison of our approach with state-of-the-art tracker. 
Observed from the visualization results, FCOT produces more precise bounding boxes and has the ability to discriminate the target between similar objects.
}
\label{fig:1}
\end{figure}

Based on the above analysis, we present a Fully Convolutional Online Tracker, termed as FCOT, to yield a conceptually simple, efficient, and more robust tracking framework.
The core contribution of FCOT is to unify the classification and regression branches into a fully convolutional architecture based on target filters, and introduce an online regression model generator (RMG) with a novel updating mechanism to produce an optimized regressor for accurate box prediction. 
In particular, we devise a simple anchor-free box regression branch to directly predict the offsets of four sides. Then, based a meta learning strategy, we optimize the regression model weights for each target with online learning to deal with target variation. 
Compared to online classification optimizer proposed in DiMP~\cite{dimp}, the difference lies in three aspects: 
\romannumeral1) a novel updating procedure is applied in regression where the online samples are used to generate dynamic model weights and then the given sample with ground-truth bounding box is served to rectify the dynamic model weights. 
In this way, it can not only introduce dynamic target information but also rectify the error accumulation.
\romannumeral2) the optimizing objective in the rectifier is different with DiMP's optimizer thus the process of the steepest descent varies. 
\romannumeral3) We only update a half of the regression model weights during online tracking which guarantees the balance between static and online models.
In addition, we present an up-sampling decoder to perform high-resolution prediction, a simple yet effective multi-scale classification branch to handle similar object confusion with a coarse-to-fine fusion and a multi-scale regression head.
Thanks to the filter-based anchor-free architecture and the online regression model generator, our FCOT is able to yield more precise tracking results, as shown in Figure~\ref{fig:1}.

Our FCOT is a general and flexible online tracking framework. In practice, the FCOT is implemented with a fully convolutional encoder-decoder architecture to keep a balance between accuracy and efficiency. Both weights of classifier and regressor on top of head branches are online optimized to be adaptive to the specific target. 
The effectiveness of our FCOT framework is demonstrated on the common tracking datasets, and it indicates our online regressor is able to consistently improve tracking performance, in particular for higher IoU criteria. 
Our main contributions are summarized as follows:
\begin{itemize}
\item We propose a unified fully convolutional architecture (FCOT) based on target filters for classification and regression branch design. 
Specifically, we devise an up-sampling decoder to perform accurate prediction, a multi-scale prediction strategy for classification branch to handle the issue of similar object confusion and a specific multi-scale regression head.
This simple tracking recipe not only allows for efficient training and deployment, but also enables online learning on both branches for accurate and robust tracking.
\item We design a Regression Model Generator (RMG), comprising a dynamic model generator and the steepest descent based rectifier, with a novel updating mechanism to online optimize the regression model. It enables the regression branch to be capable of dealing with target variation effectively while avoiding error accumulation, thus producing more precise tracking results.
\item The proposed FCOT outperforms the popular state-of-the-art trackers on seven benchmark datasets including VOT2018~\cite{vot2018}, LaSOT~\cite{lasot}, TrackingNet~\cite{trackingnet}, GOT-10k~\cite{got10k}, UAV123~\cite{uav123}, NFS~\cite{nfs}, and OTB100~\cite{otb} while running at a real-time speed.
\end{itemize} 

\section{Related Work}
Visual object tracking (VOT) is the task of tracking an arbitrary object through a video given the first-frame bounding box of the target.
Generally, visual objective tracking can be divided into target classification and regression subtasks~\cite{CRPN,Real_Time_Object_Tracking,siamfc,atom,siamrcnn}.
In this section, we first briefly review recent trackers from the two aspects. Then we cast an inspection on the online learning mechanism in tracking. Finally, we review variable object detection works related to tracking.

\paragraph{Target Classification in Tracking} 
Modern tracking methods can be categorised as siamese trackers, filter-based trackers and transformer-based trackers. The first one is based on template matching, using Siamese networks~\cite{siamfc,siamrpnPlus,Siamese_Cascaded_Region_Proposal_Networks_for_Real_Time_Visual_Tracking,StructSiam,Learning_dynamic} to perform similarity learning. Bertinetto \textit{et al.}~\cite{siamfc} first employed Siemese network to measure the similarity between the target and the search region with a tracking speeds of over 100 fps. SiamRPN~\cite{siamrpn} formulated visual tracking as a local one-shot detection task in inference by introducing a Region Proposal Network to Siamese network. SiamRPN++~\cite{siamrpnPlus} improved SiamRPN by substituting the modified AlexNet~\cite{alexnet} with Resnet-50~\cite{resnet50}, which enables the backbone to extract abundant features. 

Filter-based trackers aims at learning an adaptive filter by maximizing the response gap between target and background. Paticularly, the correlation-filter-based~\cite{kcf,eco} trackers and classifier-based trackers~\cite{mdnet} are typical methods to online update the classification model so as to distinguish the target from background. However, these approaches rely on
complicated online learning procedures that cannot be easily formulated in an end-to-end learning architecture. Bhat \textit{et al.}~\cite{dimp} and Park \textit{et al.}~\cite{meta-tracker} further learned to learn during tracking based on the meta-learning framework. DiMP~\cite{dimp} introduced a target model predictor to online optimizing the target model instructed by the discriminative loss, which achieves leading performance in various benchmarks. In our work, we employ the target model predictor to perform online classification.

Recently, transformer has been widely used~\cite{tmt,stmtrack,tt} and achieved promising achievements. These works use transformer to encode the target and the search region information simultaneously, which incorporate the target information in a pixel-to-pixel way.

\paragraph{Target Regression in Tracking} 
Previous trackers can be divided into three categories based on the task of target regression.
DCF~\cite{dcf_} and SiamFC~\cite{siamfc} employed brutal multi-scale test to estimate the target scale roughly. RPN-based trackers~\cite{siamrpn,siamrpnPlus} regressed the location shift and size difference between pre-defined anchor
boxes and target location. ATOM~\cite{atom}, DiMP~\cite{dimp} and PrDiMP~\cite{prdimp} employed an IoUNet to iteratively refine the initial multiple boxes. In this work, we take inspiration from FCOS~\cite{fcos} to regress the distance from estimated target center to the sides of the bounding box, which is similar with 
some siamese anchor-free trackers~\cite{siamfc++,siamcar,siamban,ocean,treg}. However, our FCOT is different on several important aspects. 
Above all, our FCOT is essentially a filter-based tracker with a focus on enabling online optimization for both classification and regression branches, while others are siamese trackers with fixed kernels for both branches. In addition, to fully unleash the power of FCOT, we resort to higher resolution of feature map produce classification and regression results. 

\begin{figure*}[pt]
\centering
\includegraphics[width=\linewidth]{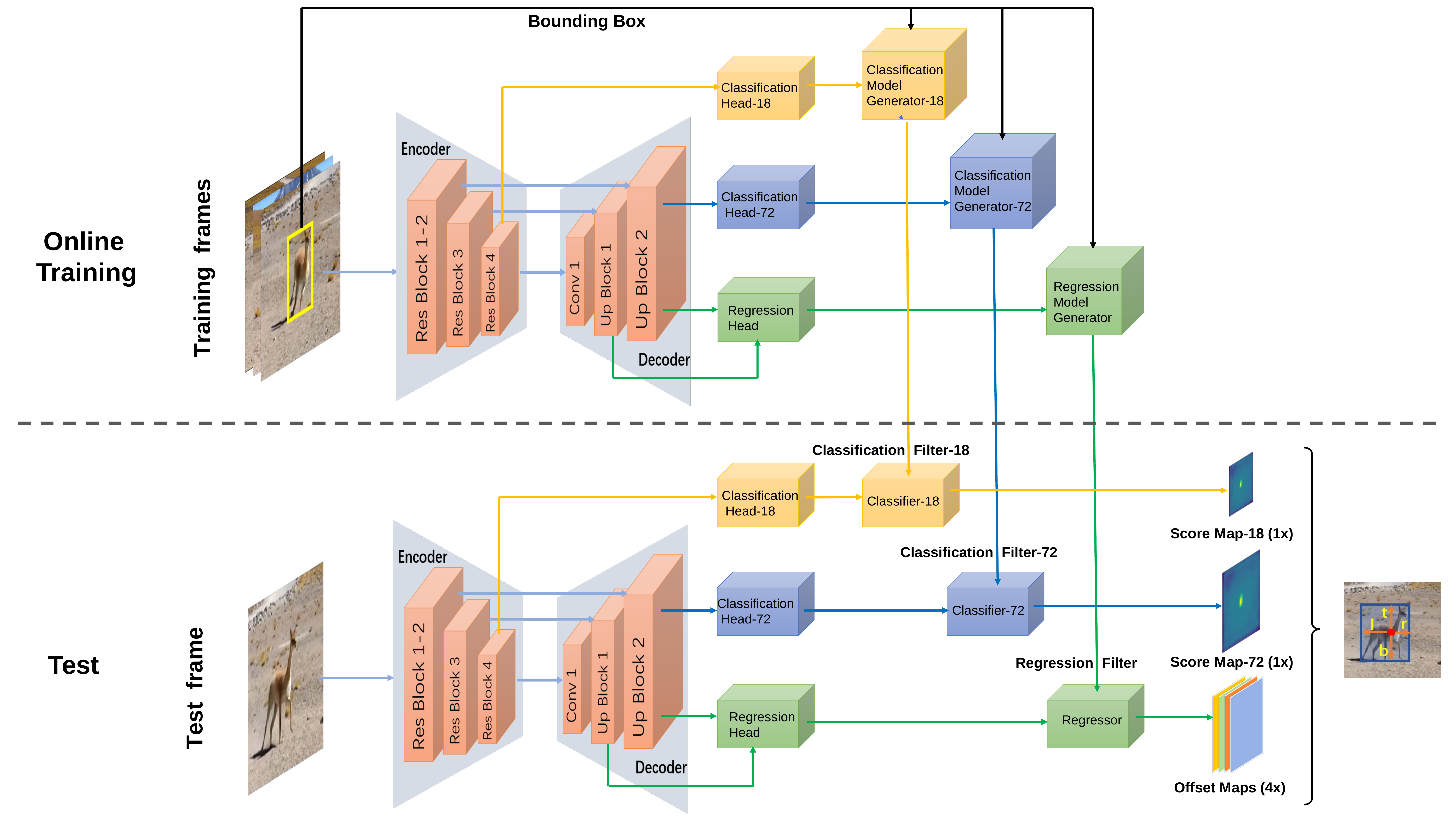}
\caption{{\bf Fully Convolutional Online Tracker}.
Our FCOT presents a fully convolutional framework for online tracking, which is composed of an encoder-decoder backbone, two classification heads and a regression head on top for task-specific feature extraction, classification and regression model generators, classifiers and regressor.
Our FCOT follows a simple online tracking recipe, where both classification and regression model generators will produce a target-specific classifier and regressor weight based on an online updated training set. These weights will be adaptive for each tracked object, thus making our FCOT more accurate and robust. Details of regression model generator could be found in Figure~\ref{fig:estimation_subnet}.
}
\label{fig:architecture}
\end{figure*}

\paragraph{Online Learning in Tracking}

Online learning possesses an important role for discriminative trackers since it can incorporate information from the background region or previous tracked frames into the model prediction. Initially, Some works~\cite{Nam_2016_CVPR,eco} propose to learn an online correlation filter or classifier to distinguish the target from background, thereby achieving impressive performance on robustness. However, they employ hand-crafted features or deep features pre-trained for object classification, which are hard formulated in an end-to-end learning framework. Then some recent works~\cite{Valmadre_2017_CVPR,6870486,Yao_2018_ECCV,Galoogahi_2017_ICCV,Park_2018_ECCV,dimp} concentrate more on formulating the classifier in trackers as an online discriminative module which can be trained end-to-end. Although these works promote the discriminant ability through introducing online learning in classification, they do not apply the strategy in regression branch. Regression branch cannot be easily optimized with online learning for each tracked target due to their complex design. Besides, there is intrinsic demands for high-quality training samples in regression. However, the predictions of online samples may be imprecise due to the error accumulation, which cause limitation to online regression.

\paragraph{Anchor-free Detection}
Tracking has a lot of relationships with object detection regardless of their unique characteristics. In detection, the anchor-based detectors~\cite{faster_rcnn} classifies pre-defined proposals called anchor as positive or negative patches, with an extra offsets regression to refine the prediction of bounding box locations. However, it introduces too much hyper-parameters (e.g. the scale/ratio of anchor boxes) and show an apparent impact on the accuracy. To overcome that, FCOS~\cite{fcos} and Centernet~\cite{centernet} formulate the object detection as anchor-free regression. They use keypoint estimation to find center points and regresses to all other object properties, such as size, 3D location, orientation, and even pose. Inspiring by that, we propose a fully convolutional anchor-free tracker to simplify the tracking pipeline and improve the accuracy. Different from FCOS~\cite{fcos}, FCOT employs target model (the dynamic convolutional filter) to incorporate target information to the test branch, thereby making it suitable for tracking. Besides, we do not use FPN since the target size are relatively stable in a search region cropped with the previous target states.

\section{Our Method}

\begin{figure*}[ht]
\centering
\includegraphics[width=12cm]{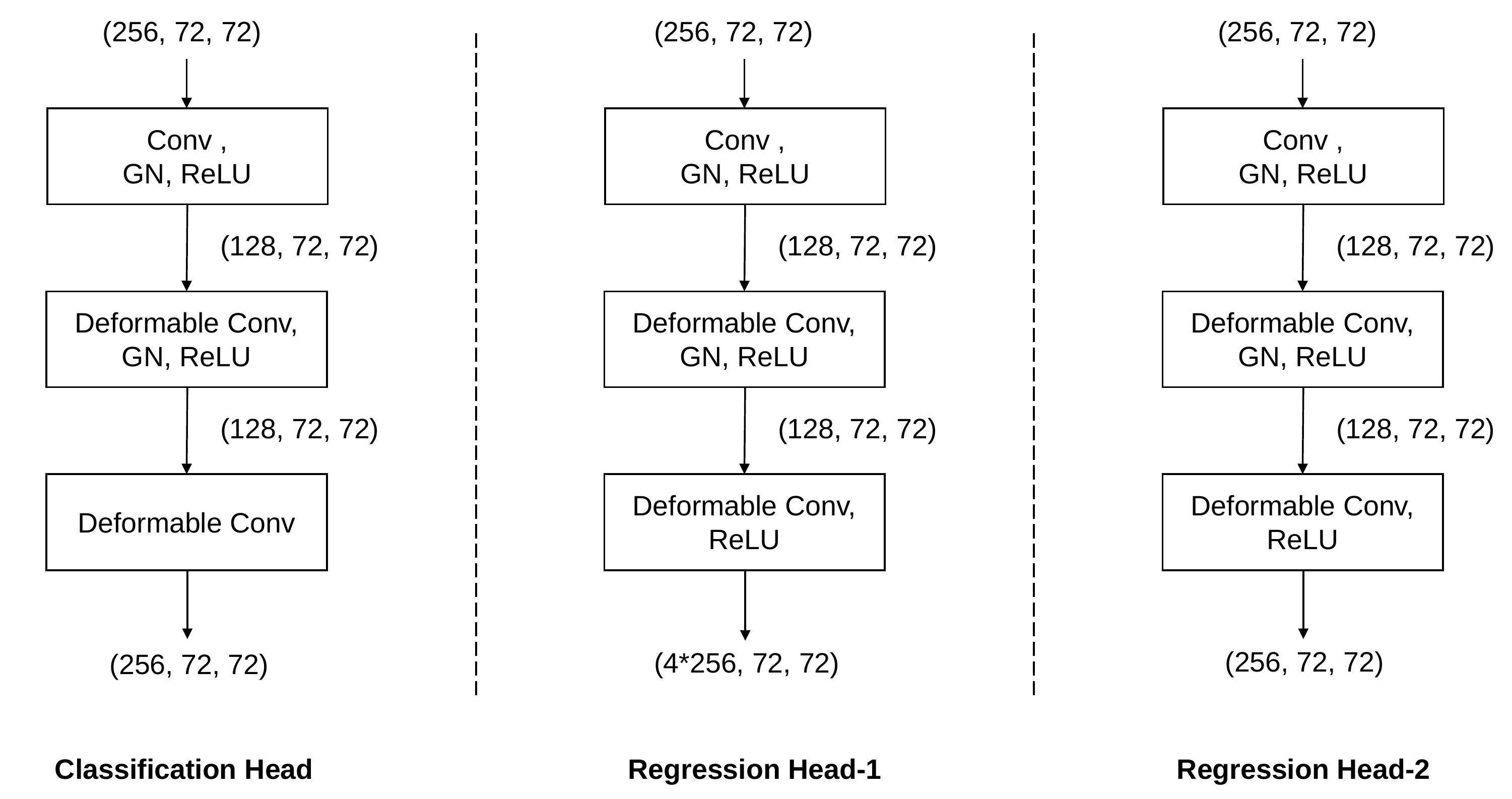}
\caption{The structure of the Classification Heads and Regression Heads, where GN represents the Group Normalization. The classification heads in training branch and test branch share the same structure and weights. The regression heads in the online training branch and the test branch are different, which are denoted as Regression Head-1 and Regression Head-2 respectively. We visualize scale-72 regression heads instead of the two-scale ones used in FCOT for simplicity. The Regression Head-1 outputs 1024 feature maps to be used for generating four regression filters while the Regression Head-2 outputs 256 feature maps to be performed a regression convolution with the four filters.}
\label{fig:architecture-heads}
\end{figure*}

In order to obtain a simple, efficient, and precise tracking method, we design a fully convolutional online tracker (FCOT), guided by the following principles.
\begin{itemize}
\item[\textbf{(\romannumeral1)}] \textbf{A unified and effective architecture.}  We hope the components of feature extraction, classification branch, and regression branch could be implemented in a single and unified network architecture. A fully convolutional network based on target filter rather than siamese framework is employed to locate the target center and regress the offsets from the target sides to the center directly, which can avoid designing hand-crafted box size estimation head such as hyper-parameters and IoU prediction in DiMP~\cite{dimp} or anchor box placement and design in SiamRPN~\cite{siamrpnPlus}. The unified fully convolutional scheme also enables the FCOT to be efficient for both training and inference.
In addition, compared with previous trackers, FCOT generates larger score map and box offset maps with an up-sampling decoder, ensuring more precise target center location and target bounding box regression.
For classification, we devise a multi-scale prediction strategy for classification branch to handle the issue of similar object confusion. For regression, we employ a carefully devised head with multi-scale feature aggregation to extract abundant information.
\item[\textbf{(\romannumeral2)}] \textbf{Accurate regression with online learning.} 
Due to the simplicity of FCOT, it's for the first time to online optimize the regression model implemented by our proposed Regression Model Generator, which comprises a dynamic model generator and the steepest descent based rectifier.
The novel updating procedure can introduce dynamic target information and avoid error accumulation. In this way, FOCT can update the regression model online thereby regressing the bounding box accurately facing the issue of target appearance changing along temporal dimension.

\end{itemize}

In this section, we describe our proposed Fully Convolutional Online Tracker (FCOT) in details. First, we describe the framework of fully convolutional tracking. Then, we introduce the regression model generator for online learning. Finally, we present the details of online tracking with the proposed FCOT.

\subsection{Fully Convolutional Tracking Framework}

In order to obtain a simple, efficient, and robust tracking method, we design a fully convolutional online tracker (FCOT).
As shown in Fig~\ref{fig:architecture}, FCOT comprises a ResNet-50~\cite{resnet50} backbone to extract general feature, classification and regression heads to generate sub-task specific feature, online model generators for the two tasks, multi-scale classifiers to locate the target center, and a convolutional regressor for estimating the offsets of the four sides. Thus, our discriminative tracker can integrate online updating target-specific information into classification and regression so as to predict bounding boxes accurately with a simple FCN-based architecture.

\paragraph{Feature extraction} 
The general features are extracted by an encoder-decoder backbone, where the encoder covers from the Layer-1 to Layer-4 of ResNet-50 and the decoder contains three convolutional layer and two simple up-sample layers. The spatial down-sampling ratio of the general feature is 4.
Then the Classification heads and the regression heads extract task-specific features to cope with classification and regression tasks separately. Specifically, we use the same classification head with DiMP~\cite{dimp} for Score Map-18. As shown in Fig~\ref{fig:architecture-heads}, the classification head for Score Map-72 and the regression head are composed of one convolutional layer, two deformable convolutional layers~\cite{deformable_conv,deformable_conv_v2}, three group normalization layers and two ReLU activation layers. 
In regression head, features extracted from the last two layers of the decoder are fused with a convolutional layer to perform accurate regression. The encoder-decoder backbone and classification branch employed for online training frames and test frames share structure and weights. While the regression heads for online dynamic generator and testing are different.

\begin{figure*}[ht]
\centering
\includegraphics[width=15cm]{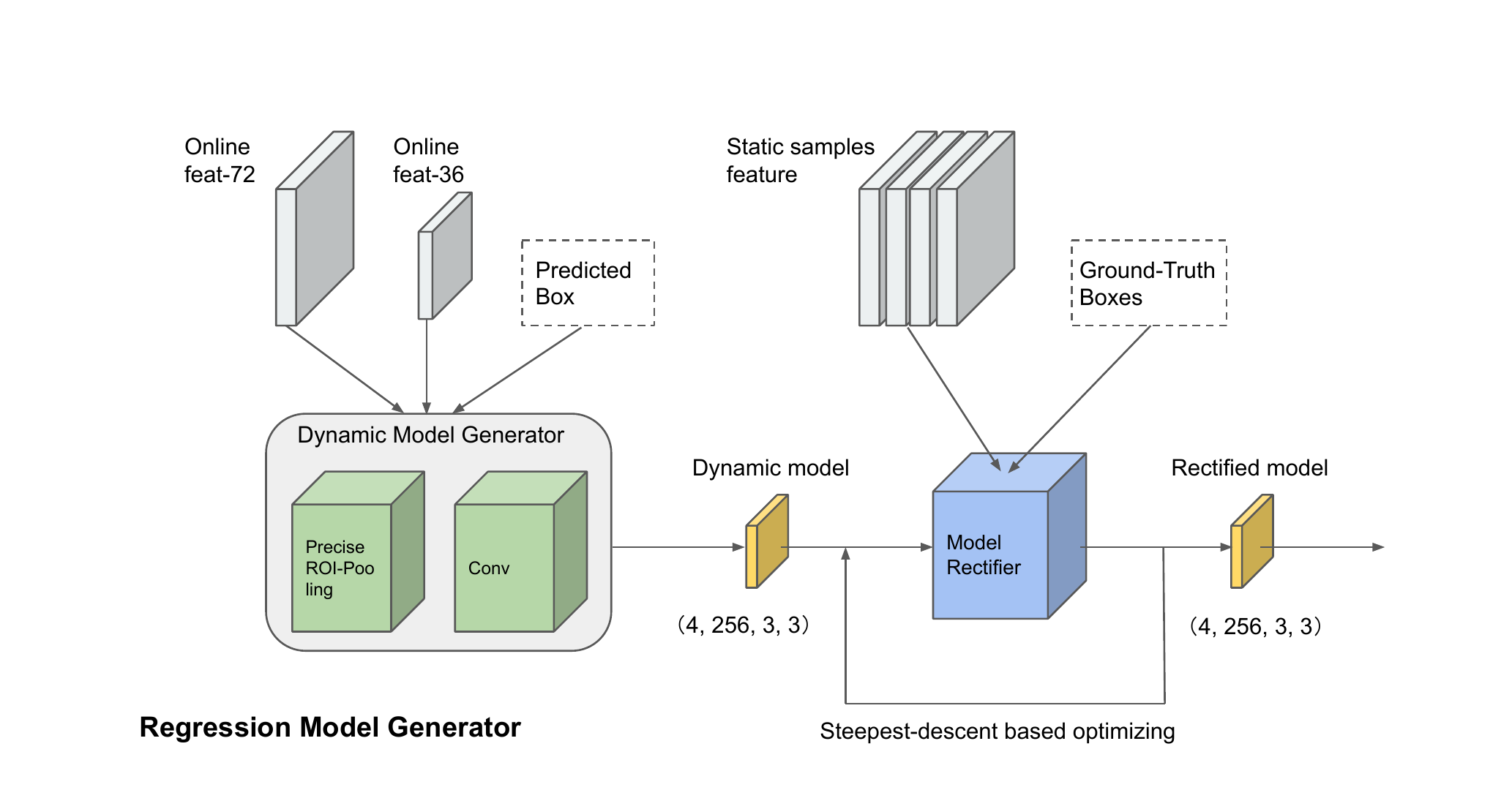}
\caption{Our proposed {\bf Regression Model Generator} produces the regressor weights in regression branch. It consists of an online dynamic model generator and a model rectifier. The dynamic model generator takes the features extracted from the regression-head-2 and bounding boxes of the online predicted samples as input to generate dynamic model. The features extracted from the regression-head-1 and bounding boxes of the given sample are then fed into the model rectifier to rectify
the regression model iteratively. Finally, the static model generated by the augmented given samples and the online model are merged.}
\label{fig:estimation_subnet}
\end{figure*}

\paragraph{Multi-scale classification}
In general, in visual tracking, score map with coarse resolution produces robust yet not accurate results, while a high resolution map is with a complementary property.
Thus, we devise a multi-scale prediction strategy for classification branch to handle the issue of similar object confusion, and also improve accuracy. As shown in Fig~\ref{fig:architecture}, Score Map-18 and Score Map-72 are generated by separate online classifiers based on the feature maps of different resolutions. Then, these two maps are fused to predict the target center. The low-resolution score map produces a rough location of the target center, while the high-resolution score map can refine the location. Without the low-resolution score map, the tracker could be easily confused by the similar objects. On the contrary, the predicted coordinates of the target center are not precise enough if only using a low-resolution score map. Specifically, the prediction of classification score maps $M_{cls}$ in our FCOT can be defined as:
\begin{equation}
\begin{split}
    M_{cls}^{(18)} = \phi^{(18)}(\omega(I_{t})) * f_{cls}^{(18)}, \\\   
    M_{cls}^{(72)} = \phi^{(72)}(\omega(I_{t})) * f_{cls}^{(72)}, \\\
    M_{cls} = \alpha M_{cls}^{(18)} + \beta M_{cls}^{(72)},
\end{split}
\end{equation}
where the parameter $\omega$ denotes the weights of an encoder-decoder backbone, $\phi^{(18)}$ is the classification head of Score Map-18 and $\phi^{(72)}$ of Score Map-72, $I_{t}$ represents for the test frame, $f_{cls}^{(18)}$ and $f_{cls}^{(72)}$ are the classification models of each branch generated by the corresponding model generators, $\alpha$ and $\beta$ are weights of the two score maps. During training, the classification target is a Gaussian function map centered at the ground-truth target center $c_{t}$.

\paragraph{Anchor-free regression}
For regression branch, we formulate the regression as a per-pixel prediction problem. We predict the offsets maps $M_{reg}$ of four sides via the regression branch, which is defined as:
\begin{equation}
    M_{reg} = \theta(\omega (I_{t})) * f_{reg}.
\end{equation}
The parameter $\theta$ is the regression head to extract regression-specific features, $f_{reg}$ represents the filter weight generated by the regression model generators and $*$ denotes a convolution operation.

For each location $(x,y)$ on the final feature maps, we can map it back onto the input image as $(\lfloor \frac{s}{2} + xs \rfloor, \lfloor \frac{s}{2} + ys \rfloor)$, where $s$ is the stride of the feature extractor(In this work, $s=4$).
Each coordinate $(x,y)$ of $M_{reg}$ is expected to a 4D vector $(l^{*},r^{*},t^{*},b^{*})$ representing the distance from $(x,y)$ to the sides of the bounding box in the final feature maps. 
Hence, the regression targets of position $(x,y)$ can be formulated as follows:
\begin{equation}
\begin{split}
    l^{*} = \lfloor \frac{s}{2} + xs \rfloor - x_{0}, \\\
    r^{*} = x_{1} -  \lfloor \frac{s}{2} + xs \rfloor, \\\
    t^{*} = \lfloor \frac{s}{2} + ys \rfloor - y_{0}, \\\
    b^{*} = y_{1} -  \lfloor \frac{s}{2} + ys \rfloor.
\end{split}
\end{equation}
During the offline training process, we regress for positions in the vicinity of the target center $c_t$ (the area with a radius of 2 in this work) rather than for the only pixel $c_t$. 

\subsection{Regression Model Generator}

Different from online classification, online regression has an intrinsic characteristic that it requires higher-quality training samples. However, the predictions of online samples may be imprecise due to the error accumulation, which cause limitation to online regression. As a consequence, the traditional online mechanism~\cite{dimp} show poor performance for regression. To alleviate the drawback, we present a novel online regression model generator comprising a dynamic model generator using online samples to incorporate online target information and a rectifier using the first frame to rectify the online model.

As shown in Fig~\ref{fig:estimation_subnet}, the regression model contains a dynamic model generator and a model rectifier. The dynamic model generator takes the features extracted from the regression-head-2 and bounding boxes of the online predicted samples as input to generate dynamic model incorporating online information which is a regression convolutional filter. The features extracted from the regression-head-1 and bounding boxes of the given sample are then input into the model rectifier to rectify the model weights iteratively. According to the design, our tracker can not only update the filter online to fit with the changing target appearance by the optimizer, but also avoid the error accumulation during online regression. 

The dynamic generator is composed of a precise ROI pooling layer~\cite{faster_rcnn} and a  convolutional layer. The dynamic generator works on the feature of the online predicted samples. And the model rectifier is derived from the following regression training loss:
\begin{equation}
    L(f) = \frac{1}{N} \sum\limits_{(X,c) \in S_{train}}\left\| \hat{M_{reg}^{(c)}}- X^{(c)} * f \right\|^2 + \left\| \eta f \right\|^2.
\end{equation}
The parameter $N$ is the length of augmented training set of the given sample $S_{train}$, $X$ is the features extracted by the regression-head-1,
$\hat{M_{reg}^{(c)}}$ denotes the distance from position $c$ to four sides of the ground-truth bounding box or the predicted box during online tracking,
$X^{(c)}$ is a portion of $X$ with an area of $3\times3$ (the same with the regression model size) centered at $c$, $f$ is the regression convolution filter, $*$ denotes convolution and $\eta$ is a regularization factor which is the only parameter to be offline trained.
The objective is to optimize the regression convolution filter $f$. 
Since the gradient descent is slow, we solve the issue with the steepest descent methodology, which compute a step length $\alpha$ to update the model as follows:
\begin{equation}
    f^{(i+1)} = f^{(i)} - \alpha \nabla{L(f^{(i)})}.
\end{equation}
The parameter $i$ denotes the number of iterations of optimizing. We compute $\alpha$ and $\nabla{L(f^{(i)})}$ according to Gauss-Newton method~\cite{numerical_optimization} as follows.

\textbf{Calculation of Step Length $\alpha$ in Equation (5):}
Similar with classification model optimizer proposed in DiMP~\cite{dimp}, we rectify the the regression model through steepest descent methodology~\cite{numerical_optimization}. 
With steepest methodology, the step length $\alpha$ can be derived as,
\begin{equation}
    \alpha = \frac { \nabla{L(f^{{i}})}^{T} \nabla{L(f^{{i}})} } 
                   {\nabla{L(f^{{i}})}^{T}  H^{(i)}  \nabla{L(f^{{i}})} },
\end{equation}
where $\nabla{L(f^{{i}})}$ is the gradient of the loss with respect to the filter $f^{(i)}$, $H^{(i)}$ is the Hessian of the loss (1). According to Gauss-Newton method~\cite{numerical_optimization}, we can approximate the Hessian $H^{(i)}$ with $(J^{(i)})^{T}J^{(i)}$ for our least-squares formulation (1), where $J^{(i)}$ is the Jacobian of the rasiduals at $f^{(i)}$~\cite{dimp}. Thus, $\alpha$ can be denoted as,
\begin{align}
    \alpha &\approx \frac { \nabla{L(f^{{i}})}^{T} \nabla{L(f^{{i}})} } 
        {\nabla{L(f^{{i}})}^{T}  (J^{(i)})^{T} J^{(i)}  \nabla{L(f^{{i}})}}\\
           &= \frac{\left\| \nabla{L(f^{{i}})} \right\|^2}   
             {\left\| J^{(i)}  \nabla{L(f^{{i}})} \right\|^2}.
\end{align}
Then the gradient $\nabla{L(f)}$ of the loss (1) with respect to the regression model $f$ can be computed as,
\begin{equation}
    \nabla{L(f)} = \frac{2}{N} \sum\limits_{(X,c) \in S_{train}} (X^{(c)})^T * (\hat{M_{reg}^{(c)}}- X^{(c)} * f) + 2\eta ^2 f.
\end{equation}
For similarity, we use $\left\| h \right\|^2$ to represent for the other term $\left\| J^{(i)}  \nabla{L(f^{{i}})} \right\|^2$. Similar with the classification model~\cite{dimp}, we can derive that,
\begin{equation}
    \left\| h \right\|^2 = \frac{1}{N} \sum\limits_{(X,c) \in S_{train}} 
                           \left\| X^{(c)} * \nabla{L(f^{{i}})} \right\|^2 +
                           \left\| \eta \nabla{L(f^{{i}})} \right\|^2.
\end{equation}
Consequently, $\alpha$ can be easily computed in PyTorch based on the euquation (5) and (6).

\textbf{Discussion with DiMP's model predictor:} Comparing the RMG with the model predictor proposed in DiMP~\cite{dimp}, the main differences lie in three aspects.
First, the overall updating paradigm is different. For RMG, a dynamic model generator is applied to introduce online target information and then the model rectifier adjust the model weights based on the fist given sample. It depends on the unique characteristic of the online regression that high-quality samples are required to avoid error accumulation. Second, the network between the rectifier in RMG and optimizer in DiMP varies as they are derived from different optimization objectives. The rectifier only casts supervision on the target center while the other supervises the whole target so as to distinguish the target between background. Third, there are some parameters of the discriminative loss in DiMP's optimizer that are learned during offline training, leading to heavy training cost. However, the only penalty coefficient $\eta$ in the rectifier of our RMG is needed to be trained offline, which can be easily optimized with less epochs.

\subsection{Online Tracking}\label{online_tracking}
After introducing the FCOT framework and RMG, we are ready to present the online tracking recipe. We perform data augmentation to the first frame with translation, rotation,and blurring, yielding a total of 23 initial training samples.
For classification and regression, the target models are first generated by the model predictor with the augmented training samples. We denote it as the static model. The static models are generated in two steps, which are using the initializer and the steepest descent based optimizer respectively. Then the static classification model are updated using the online training set during online tracking procedure. Specifically, we present a simple strategy to update classification model with online learning, which adding the frames with the highest classification score every $n$ frames to the online training set so as to keep a high quality of the classification training samples. 
While for online regression, online model is produced as described in the previous section every $n$ frames.
Then we merge the online model $f_{on}$ with the static model $f_{st}$, so the current model $f_{cur}$ can be formulated as:
\begin{equation}
    f_{cur} = \lambda f_{on} + (1-\lambda) f_{st}.
\end{equation}
These strategies are proved to be effective to improve the accuracy and robustness of FCOT.

\section{Experiments}

\subsection{Implementation Details}

Our FCOT is implemented with Pytorch based on the project Pytracking~\cite{pytracking}. 
We use ADAM~\cite{adam} with learning rate decay of 0.2 at the epoch of 25 and 45. 
Our offline training are performed at two stages. First, we train the entire network for 100 epochs except for the regression rectifier in the model generator. Then we train the regression rectifier with the rest of the network freezed for 5 epochs. The total loss for offline training can be formulated as $L_{tot}=\gamma L_{cls} + \delta L_{reg}$, which $\gamma$ and $\delta$ are hyper-parameters. For classification branch, we use the same loss $L_{cls}$ and training strategies as DiMP~\cite{dimp}. 
For regression, We use IoU loss~\cite{iounet} as $L_{reg}$.
We spend around 50 hours training the whole model on 8 RTX 2080ti GPUs.
While for GOT-10k test, we train our tracker by only using the GOT10k train split following its standard protocol. 
The regression model is a convolution kernel with the size of $((4\times256)\times3\times3)$ and the two classification models with the size of $(256\times4\times4)$.
For inference, the average tracking speed is over 40 FPS (testing on got10k~\cite{got10k} dataset) on a single RTX 2080ti GPU. 

\subsection{Exploration Study}
Here, we perform an extensive analysis on the proposed online anchor-free tracker. Experiments are performed on VOT2018 dataset evaluated on the accuracy, robustness and EAO metrics. For fair comparison, we always report the second highest EAO scores with fair hyper-parameters tuning. 

\paragraph{Analysis of multi-scale regression head}
The first line in Table~\ref{ablation_framework} is the baseline which comprises the resnet-50 backbone, the same classification branch as in DiMP and the scale-18 regression branch without online model generator.
Comparing the experiments on the second line with the baseline, we find that the setting of using multi-scale feature performs better than just using single-scale feature, which increases the performance by 0.008 for EAO and 0.006 for accuracy. These results indicate the effectiveness of multi-scale regression head which can boost the precision of the bounding box with abundant feature.  
\paragraph{Analysis of multi-scale classification and up-sampling decoder}
To verify the performance of multi-scale classification and the up-sampling decoder, we conduct the experiment as shown on the third line and the fourth line in Table~\ref{ablation_framework}. 
\begin{table}[pt]
\fontsize{9}{11}\selectfont 
  \caption{Ablation analysis of FCOT framework on the VOT2018 data set. Reg-2S represents for multi-scale regression head, Cls-2S for multi-scale classification, UP for UP-Sample block and Online-Reg for online regression model generator. The best results are highlighted by {\bf bold}.}
  \label{ablation_framework}
  \centering
  \begin{tabular}{cccc|ccc}
    \toprule
     Reg-2S & Cls-2S & UP & Online-Reg & Acc. & Rob. & EAO  \\
    \midrule
        && & & 0.547  & 0.126  & 0.443  \\
        \checkmark& & & & 0.553  & \bf 0.112  & 0.451  \\
        \checkmark& \checkmark& & & 0.553  & \bf 0.112  &  0.447 \\
      \checkmark& \checkmark&\checkmark& &  0.597 &  0.122 & 0.463 \\
      \checkmark& \checkmark&\checkmark& \checkmark& \bf 0.612 & \bf 0.112 & \bf 0.491 \\
    \bottomrule
  \end{tabular}
\end{table}

\begin{table}[pt]
\fontsize{9}{11}\selectfont 
  \caption{Ablation analysis of online regression model generator on the VOT2018 data set. The best results are highlighted by {\bf bold}.}
  \label{ablation_online}
  \centering
  \begin{tabular}{ccc|ccc}
    \toprule
     Init-Filter & Init-Rect & Online & Acc. & Rob. & EAO  \\
    \midrule
        \checkmark& & & 0.597  & 0.122  & 0.463  \\
        \checkmark& \checkmark& & 0.605  & 0.131  & 0.466  \\
        & &\checkmark &  0.599 &  0.126 & 0.442 \\
      \checkmark& \checkmark & \checkmark& \bf 0.612 & \bf 0.112 & \bf 0.491 \\
    \bottomrule
  \end{tabular}
\end{table}
Compared with the second line, the performance of using two-scale score maps from layer-3 and layer-4 of resnet-50 respectively has no improvement. Furthermore, the multi-scale classification with adding up-sampling decoder, by which the scale of one of the score map is increased to 72, improves the accuracy by 0.044 and the EAO by 0.012. It proves that high-resolution and low-resolution score maps contribute to the accuracy and robustness respectively and up-sampling decoder is important in the framework.
It can demonstrate that multi-scale prediction strategy with up-sampling for classification is helpful to improve both accuracy and robustness of FCOT.
\paragraph{Analysis of the online regression model generator} 
To explore the effect of online learning for regression on both accuracy and robustness, we conduct experiments as on the last line in Table~\ref{ablation_framework}. Compared with the experiment on the fourth line, adding the online model generator for regression branch can increase tracking performance by 0.015 of accuracy and 0.028 of EAO on VOT2018. It demonstrates that the proposed online regression model generator can produce more precise box and thus increase the tracking performance. More detailed analysis 
of the online model generator is conducted on the following section.

\begin{table}[pt]
\fontsize{9}{11}\selectfont 
  \caption{Analysis of the online updating mechanism of regression model on the VOT2018 data set. We compare ours method (Ours) with the traditional online updating mechanism (Trad) employed in DiMP~\cite{dimp}. The best results are highlighted by {\bf bold}.}
  \label{comparison_rmg}
  \centering
  \begin{tabular}{cc|ccc}
    \toprule
     $\lambda_{reg}$ & Type &  Accuracy & Robustness & EAO  \\
     \midrule
    0.0 & Neither & 0.605  & 0.131  & 0.466 \\
    \midrule
    \multirow{2}{*}{0.2} & Trad & 0.603  & 0.117  & 0.464 \\
    & Ours & 0.604  & 0.131  & 0.469 \\
    \midrule
    \multirow{2}{*}{0.4} & Trad & 0.604  & 0.136  & 0.455 \\
    & Ours & 0.610  & 0.131  & 0.472 \\
    \midrule
    \multirow{2}{*}{0.6} & Trad & 0.589  & 0.150  & 0.438 \\
    & Ours & \bf 0.612  & \bf 0.112  & \bf{0.491} \\
    \midrule
    \multirow{2}{*}{0.8} & Trad & 0.574  & 0.145  & 0.442 \\
    & Ours & 0.604  & 0.117  & 0.470 \\
    \midrule
    \multirow{2}{*}{1.0} & Trad & 0.545  & 0.164  & 0.415 \\
    & Ours & 0.599  & 0.126  & 0.442 \\
    \bottomrule
  \end{tabular}
\end{table}

\begin{table}[pt]
\fontsize{9}{11}\selectfont  
  \caption{Ablation study on the fusion rate $\lambda_{reg}$ of initial regression model and online regression model on the VOT2018 data set. The best results are highlighted by {\bf bold}.}
  \label{ablation_fusion_rate}
  \centering
  \begin{tabular}{c|ccc}
    \toprule
     $\lambda_{reg}$  &  Accuracy & Robustness & EAO  \\
     \midrule
    0.0  & 0.605  & 0.131  & 0.466 \\
    0.1 &  0.608 & 0.122  & 0.481 \\
    0.2 & 0.604  & 0.131  & 0.469 \\
    0.3 &  0.604 & 0.117  & 0.472 \\
    0.4 & 0.610  & 0.131  & 0.472 \\
    0.5 & 0.611  & 0.131  & 0.459 \\
    0.6 & \bf 0.612  & \bf 0.112  & \bf{0.491} \\
    0.7 & 0.608  &  0.131 & 0.452 \\
    0.8 & 0.604  & 0.117  & 0.470 \\
    0.9 & 0.608 & 0.140  & 0.442 \\
    1.0 & 0.599  & 0.126  & 0.442 \\
    \bottomrule
  \end{tabular}
\end{table}

\begin{table*}[pt]
\begin{center}
\fontsize{9}{11}\selectfont  
\caption{State-of-the-art comparison on the VOT2018 data set in terms of Accuracy, Robustness and EAO. The best two results are highlighted in \textcolor{red}{\bf RED} and \textcolor{blue}{\bf BLUE} fonts.} 
\label{table:vot}
\setlength{\tabcolsep}{1.3mm}{
\begin{tabular}{c c c c c c c c c c c c c}
\toprule
               &SATIN & UPDT & CRPN & STRCF &ATOM&SiamRPN++&DiMP-50&SiamFC++&SiamBAN& Ocean & CGACD &FCOT\cr
                &~\cite{satin}  &~\cite{updt}
                &~\cite{CRPN}   &~\cite{strcf}
                   &~\cite{atom}   &~\cite{siamrpnPlus}& ~\cite{dimp}&~\cite{siamfc++}  &~\cite{siamban}       &~\cite{ocean} &~\cite{CGACD} & \cr
\midrule
    Accuracy & 0.490 & 0.536 & 0.550 & 0.523  &0.590&0.600&0.597&\textcolor{red}{\bf0.618}&\textcolor{blue}{\bf0.615}&0.592& 0.615 & 0.612\cr
    Robustness & - & 0.184 & 0.320 & 0.215   &0.204&0.234&0.153&0.165&0.172&\textcolor{blue}{\bf0.117}& 0.173& \textcolor{red}{\bf 0.112}\cr
    EAO  & 0.282 & 0.378 & 0.273 & 0.345  &0.401&0.414&0.440&0.442&0.449&\textcolor{blue}{\bf0.489} & 0.449 &\textcolor{red}{\bf 0.491}\cr
\bottomrule
\end{tabular}}
\end{center}
\end{table*}

\begin{table*}[pt]
\begin{center}
\fontsize{9}{11}\selectfont  
\caption{State-of-the-art comparison on the GOT-10k test set in terms of average overlap (AO), and success rates (SR) at overlap thresholds 0.5 and 0.75. The best two results are highlighted in \textcolor{red}{\bf RED} and \textcolor{blue}{\bf BLUE} fonts.}
\label{table:got}
\setlength{\tabcolsep}{1.3mm}{
\begin{tabular}{cccccccccccccc}
\toprule\noalign{\smallskip}
 \multirow{2}{*}{} & KCF & SRDCF & Staple & BACF & MDNet & ECO & SiamFC & ATOM & DiMP-50 & SiamFC++ & SiamCAR & OCEAN & FCOT \\   
 & \cite{kcf} & \cite{srdcf} & \cite{staple} & \cite{bacf}
 & \cite{mdnet} & \cite{eco} & \cite{siamfc} & \cite{atom} &\cite{dimp} & \cite{siamfc++} & \cite{siamcar} & \cite{ocean}& \\
\noalign{\smallskip}
\midrule
\noalign{\smallskip}
SR$_{0.5}$($\%$) & 17.7 & 22.7 & 23.9 & 26.2 & 30.3 & 30.9 &35.3 & 63.4 & \bf {\color{blue} 71.7} & 69.5 & 67.0 & 72.1 & \bf {\color{red} 76.6} \\
SR$_{0.75}$($\%$) & 6.5 & 9.4 & 8.9 & 10.1 & 9.9 & 11.1 &9.8 & 40.2 & \bf {\color{blue} 49.2} & 47.9 & 41.5 & - & \bf {\color{red} 52.1} \\
AO($\%$) & 20.3 & 23.6 & 24.6 & 26.0 & 29.9 & 31.6 & 34.8 & 55.6 & \bf {\color{blue} 61.1} & 59.5 & 56.9 & 61.1 & \bf {\color{red} 63.4} \\
\bottomrule
\end{tabular}}
\end{center}
\end{table*}

\paragraph{Analysis of online regression}
Firstly, we conduct thorough analysis on the online regression model. In table~\ref{ablation_online}, \textit{Init-Filter} denotes only using the given first sample to generate initial filter with the dynamic model generator, \textit{Init-Rect} represents for adding the rectifier with the augmented training samples of the given one as input and \textit{Online} denotes employing the proposed RMG for online regression. It can be seen that adding \textit{Init-Rect} can improve the accuracy slightly. Besides, the performance of only using the online model is not so good as the static model. However, when we merge the the online model to the static model (generated by \textit{Init-Filter} and \textit{Init-Rect}), the performance improves to the highest EAO of 0.491 and accuracy of 0.612. We can derived that introducing dynamic information by the proposed RMG can tackle the issue of target appearance changing along temporal dimension, thereby producing more accurate prediction.

Furthermore, we compare the online updating mechanism of RMG with that similar as DiMP (\textit{Trad}). We can see from Table~\ref{comparison_rmg} that, \textit{Trad} method gets the best EAO of 0.464 which below than that without online regression of 0.466. While ours gets the highest EAO of 0.491 with the fusion rate of 0.6. It indicates that the proposed dynamic model generator introduce the online target information and the rectifier can avoid error accumulation to a certain extent so as to improve the tracking performance.

\paragraph{Analysis of the static model and online model fusion strategy}
Here, we analyze the impact of the fusion rate of the static model generated by the augmented ground-truth samples and the online model generated by the predicted samples for regression branch. The experiments are shown in Table~\ref{ablation_fusion_rate}. When only use the static regression model, the EAO is 0.466 and the accuracy is 0.605. When only use the online regression model, the EAO and accuracy is 0.442 and 0.599 respectively, which are lower that the former one. When the fusion rate is 0.6, it has the best performance, which proves the effectiveness of the model fusion strategy.

\begin{figure*}[pt]
\centering
\includegraphics[width=14cm]{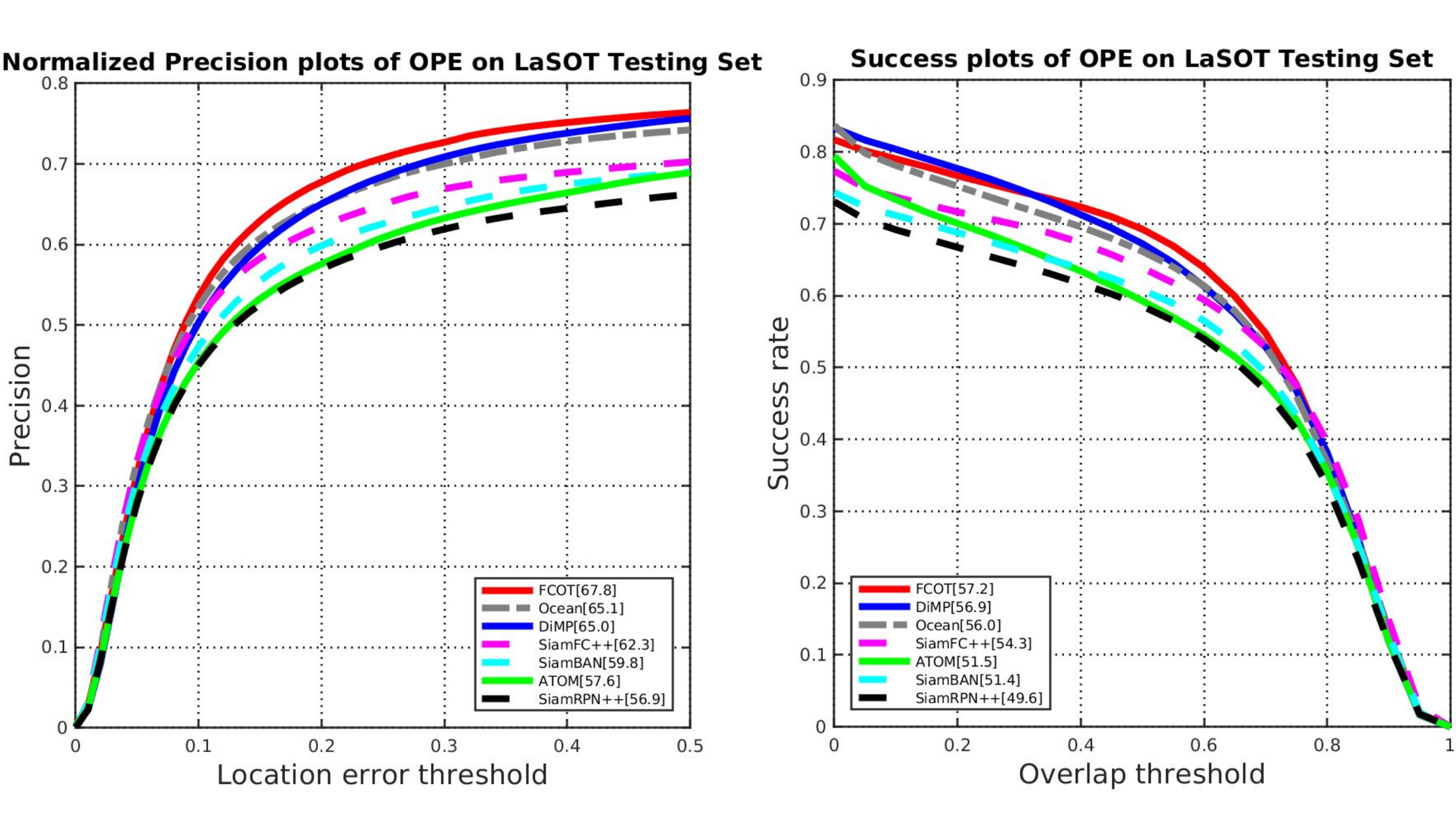}
\caption{State-of-the-art comparison on the LaSOT dataset. Best viewed with zooming in.}
\vspace{-5mm}
\label{fig:lasot}
\end{figure*}

\setlength{\tabcolsep}{4pt}
\begin{table*}[pt]
\fontsize{9}{11}\selectfont
\begin{center}
\caption{State-of-the-art comparison on the TrackingNet~\cite{trackingnet} dataset in terms of success rate, precision and normalized precision. The best two results are highlighted in \textcolor{red}{\bf RED} and \textcolor{blue}{\bf BLUE} fonts.}
\label{table:tn}
\setlength{\tabcolsep}{1.15mm}{
\begin{tabular}{ccccccccccccc}
\toprule\noalign{\smallskip}
\multirow{2}{*}{}  & ECO & SiamFC & GFS-DCF & DaSiamRPN & CRPN & SPM-Tracker & MDNet &ATOM &SiamRPN++ & DiMP & SiamFC++  & FCOT \\
   & \cite{eco} & \cite{siamfc} & \cite{gfs_dcf} & \cite{dasiamrpn} & \cite{CRPN} & \cite{spm_tracker} & \cite{mdnet} & \cite{atom} & \cite{siamrpnPlus} & \cite{dimp} &\cite{siamfc++}  & \\
\noalign{\smallskip}
\midrule
\noalign{\smallskip}
Prec.($\%$) &49.2 & 53.3 & 56.6 & 59.1 & 61.9 & 66.1 & 56.5 & 64.8 & 69.4 & 68.7 & \bf{\color{blue}70.5} & \bf {\color{red} 72.6} \\
NP.($\%$)&61.8 & 66.6 & 71.8 & 73.3 & 74.6 & 77.8 & 70.5& 77.1 & 80.0 & \bf{\color{blue} 80.1} & 80.0 & \bf {\color{red} 82.9} \\
Succ.($\%$)&55.4 &57.1 & 60.9 & 63.8 & 66.9 & 71.2 & 60.6 & 70.3 & 73.3  & \bf{\color{blue}74.0} & \bf{\color{red} 75.4} & \bf {\color{red} 75.4} \\
\bottomrule
\end{tabular}}
\end{center}
\end{table*}
\begin{table*}[pt]
\begin{center}
\fontsize{9}{11}\selectfont
\caption{State-of-the-art comparison on the UAV123 dataset in terms of success rate (AUC) and precision.The best two results are highlighted in \textcolor{red}{\bf RED} and \textcolor{blue}{\bf BLUE} fonts.}
\label{table:uav}
\setlength{\tabcolsep}{1.35mm}{
\begin{tabular}{ccccccccccccc}
\toprule\noalign{\smallskip}
 \multirow{2}{*}{} & MEEM & Staple & SRDCF & SiamFC & ECO & MDNet & DaSiamRPN &  SiamRPN & SiamRPN++ & ATOM & DiMP-50 & FCOT \\
   & \cite{meem} & \cite{staple} & \cite{srdcf} & \cite{siamfc}
   & \cite{eco} & \cite{mdnet} & \cite{dasiamrpn} & \cite{siamrpn} & \cite{siamrpnPlus} & \cite{atom} & \cite{dimp}  & \\
\noalign{\smallskip}
\midrule
\noalign{\smallskip}
AUC($\%$) & - & 45.0 & 46.3 & 48.5 & 52.5 & 52.8 & 56.9 & 52.7 & 61.3 & 63.2 & \bf {\color{blue} 64.3} & \bf {\color{red} 65.6} \\
Prec.($\%$) & 57.0 & 61.4 & 62.7 & 64.8 & 74.1 & - & 72.4 & 74.8 & 80.7 & 84.4 & \bf {\color{blue} 84.9} & \bf {\color{red} 87.3} \\
\bottomrule
\end{tabular}}
\end{center}
\end{table*}
\begin{table*}[pt]
\begin{center}
\fontsize{9}{11}\selectfont
\caption{State-of-the-art comparison on the NFS~\cite{nfs} dataset in terms of success rate (AUC) and precision. The best two results are highlighted in \textcolor{red}{\bf RED} and \textcolor{blue}{\bf BLUE} fonts.}
\label{table:nfs}
\setlength{\tabcolsep}{1.6mm}{
\begin{tabular}{cccccccccccccc}
\toprule\noalign{\smallskip}
\multirow{2}{*}{} &BACF &DSST& HCF & HDT & SAMF& GOTURN & CCOT &ECO & MDNet & UPDT & ATOM & DiMP-50 & FCOT \\
   &\cite{bacf}&\cite{dsst}& \cite{hcf}& \cite{hdt}& \cite{samf}&\cite{goturn}
   & \cite{ccot} & \cite{eco} & \cite{mdnet} & \cite{updt} & \cite{atom} &\cite{dimp}  & \\
\noalign{\smallskip}
\midrule
\noalign{\smallskip}
AUC ($\%$) &34.0& 28.0 & 29.5 & 40.3 & 29.2 & 33.4 & 48.8 & 46.6 & 42.2 & 53.7 & 58.0 & \bf {\color{blue}61.5} & \bf {\color{red}62.7} \\
Precision ($\%$) &- & -& -& -& -&- & - & - & - & - & 70.0 & \bf {\color{blue}74.1} & \bf {\color{red}75.4}\\
\bottomrule
\end{tabular}}
\end{center}
\end{table*}

\begin{figure*}[pt]
\centering
\includegraphics[width=14cm]{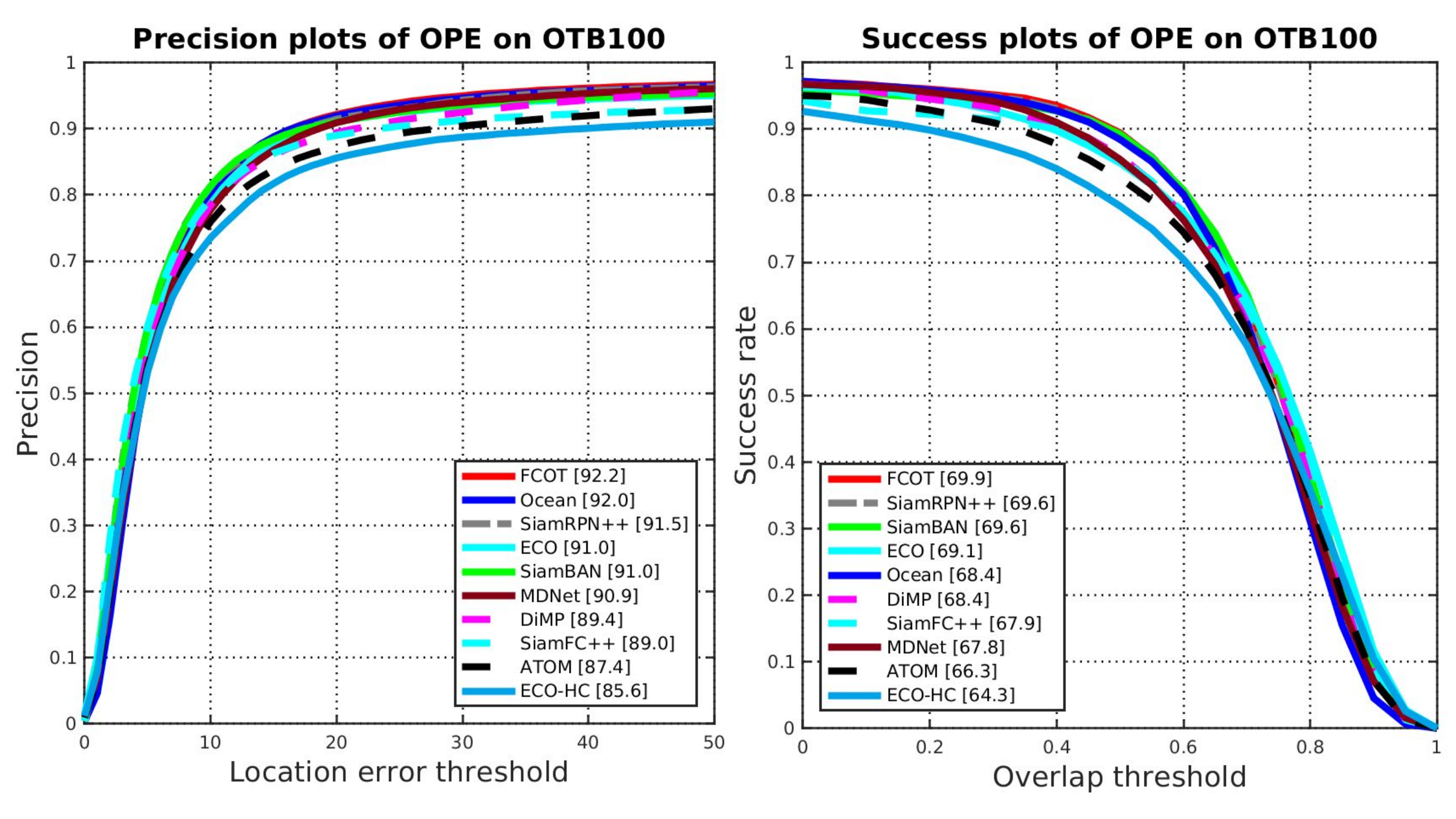}
\caption{State-of-the-art comparison on the OTB100 dataset. Best viewed with zooming in.}
\label{fig:otb}
\end{figure*}

\subsection{Quantitative Analysis}

We test the proposed FCOT on seven tracking benchmarks, including VOT2018~\cite{vot2018}, GOT10k~\cite{got10k}, LaSOT~\cite{lasot}, TrackingNet~\cite{trackingnet},UAV123~\cite{uav123}, OTB100~\cite{otb} and NFS~\cite{nfs}, and compare our results with anchor-free trackers and some state-of-the-art trackers.

\paragraph{VOT2018}
VOT2018 consists of 60 videos with several challenges including fast motion, occlusion, etc.
And the performance is evaluated in terms of robustness (failure rate) and accuracy (average overlap in the course of successful tracking). The two measures are merged in a single metric, Expected Average Overlap (EAO), which provides the overall performance ranking.
Our FCOT are tested on the dataset in comparison with state-of-the-art trackers and other anchor-free trackers.
As shown in Table~\ref{table:vot}, FCOT achieves the top-ranked performance on EAO criteria of 0.491 and Robustness of 0.112, which outperforms the previous tracker DiMP with a large margin of 5.1\% of EAO and surpass other anchor-free trackers.
This suggests that our fully convolutional online tracker can generate precise bounding boxes and so that improve the tracker's Discriminant performance.

\paragraph{GOT10k}

GOT10k~\cite{got10k} is a large-scale dataset with over 10000 video segments and has 180 segments for the test set. Apart from generic classes of moving objects and motion patterns, the object classes in the train and test set are zero-overlapped. 
We show state-of-the-art comparison on Table~\ref{table:got}. DiMP achieves an average overlap(AO) score of 61.1\%. Compared with DiMP, Our FCOT improves 2.3\% of AO, 4.9\% and 2.9\% of the success rate of threshold 0.5 and 0.75 respectively, which demonstrates the ability of FCOT to produce robust location and accurate bounding boxes.

\paragraph{LaSOT}

LaSOT~\cite{lasot} has 280 videos in its test set. With an average of 2500 frames, sequences of LaSOT are longer than other dataset, which poses great challenges to trackers. We evaluate our FCOT on the test set to validate its long-term capability. The results are shown on Fig~\ref{fig:lasot}. FCOT reaches the highest normalized precision of 67.8\%, increasing the previous method DiMP by 2.8\%. This shows our tracker can localize the target precisely. FCOT achieves a success score of 57.2\%, surpass other anchor-free trackers including OCEAN~\cite{ocean}, SiamFC++~\cite{siamfc++} and SiamBAN~\cite{siamban}. 
It's remarkable that FCOT performs better than DiMP under the high-iou criterion, which proves that online learning are effective in improving the precision of the bounding box regression. 

\paragraph{TrackingNet}
 
TrackingNet~\cite{trackingnet} provides more than 30K videos with more than 14 million dense bounding box annotations,which consists of 511 videos, with an average of 441 frames per sequence. The dataset covers a wide selection of object classes in broad and diverse context.
We validate FCOT on its test set. As shown in Table~\ref{table:tn}, our FCOT reaches a precision score of 72.6\% and a normalized score of 82.9\% surpassing DiMP and SiamFC++ by over 2\%. It proves that FCOT achieve leading performance in large-scale datasets among anchor-free trackers.

\begin{figure*}[pt]
\centering
\includegraphics[width=\linewidth]{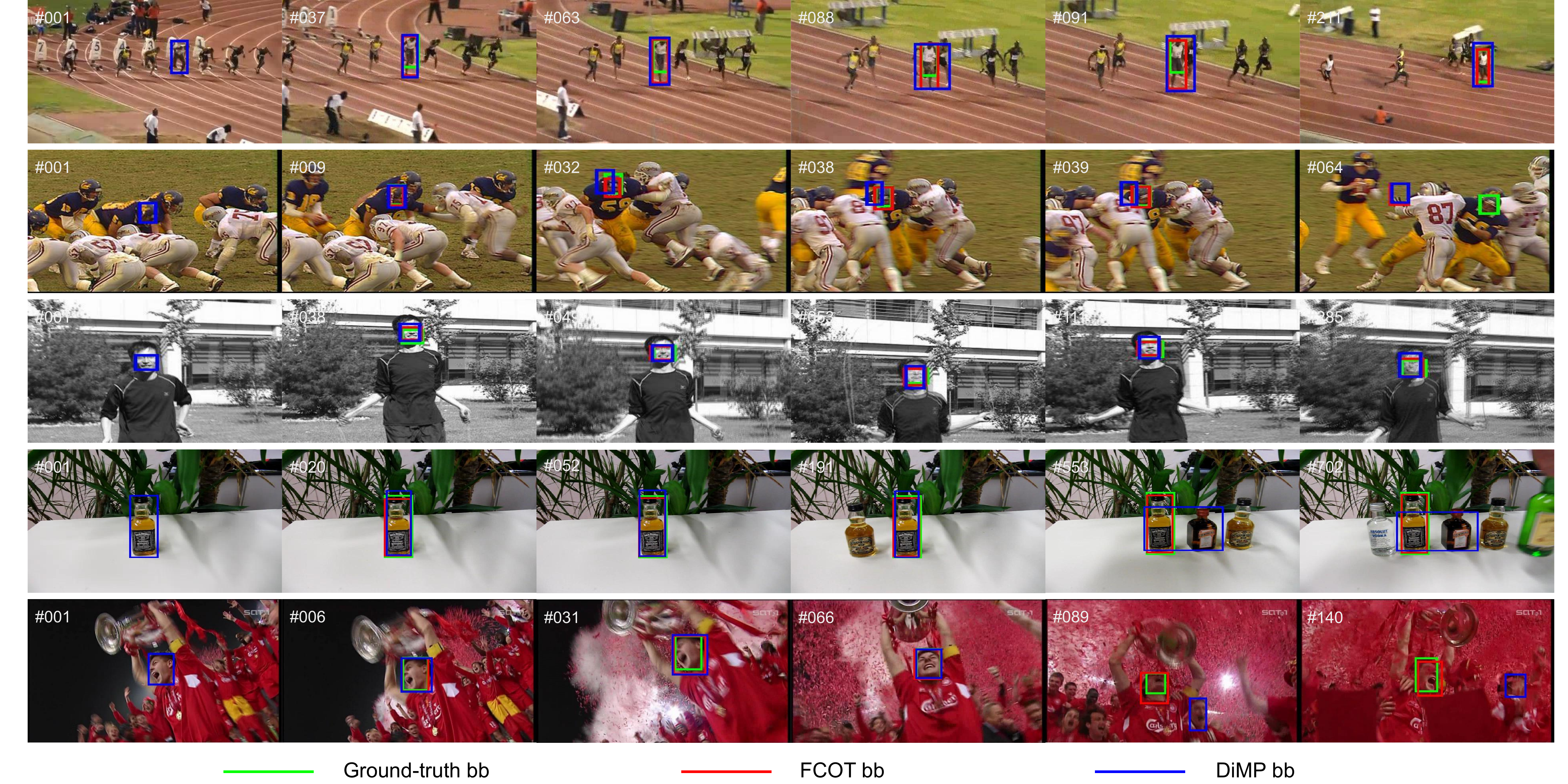}
\caption{Visualization results of DiMP and our FCOT on OTB100 dataset~\cite{otb}. 
Best viewed with zooming in.}
\label{fig:sup1}
\vspace{-5mm}
\end{figure*}

\begin{figure}[pt]
\centering
\includegraphics[width=8cm]{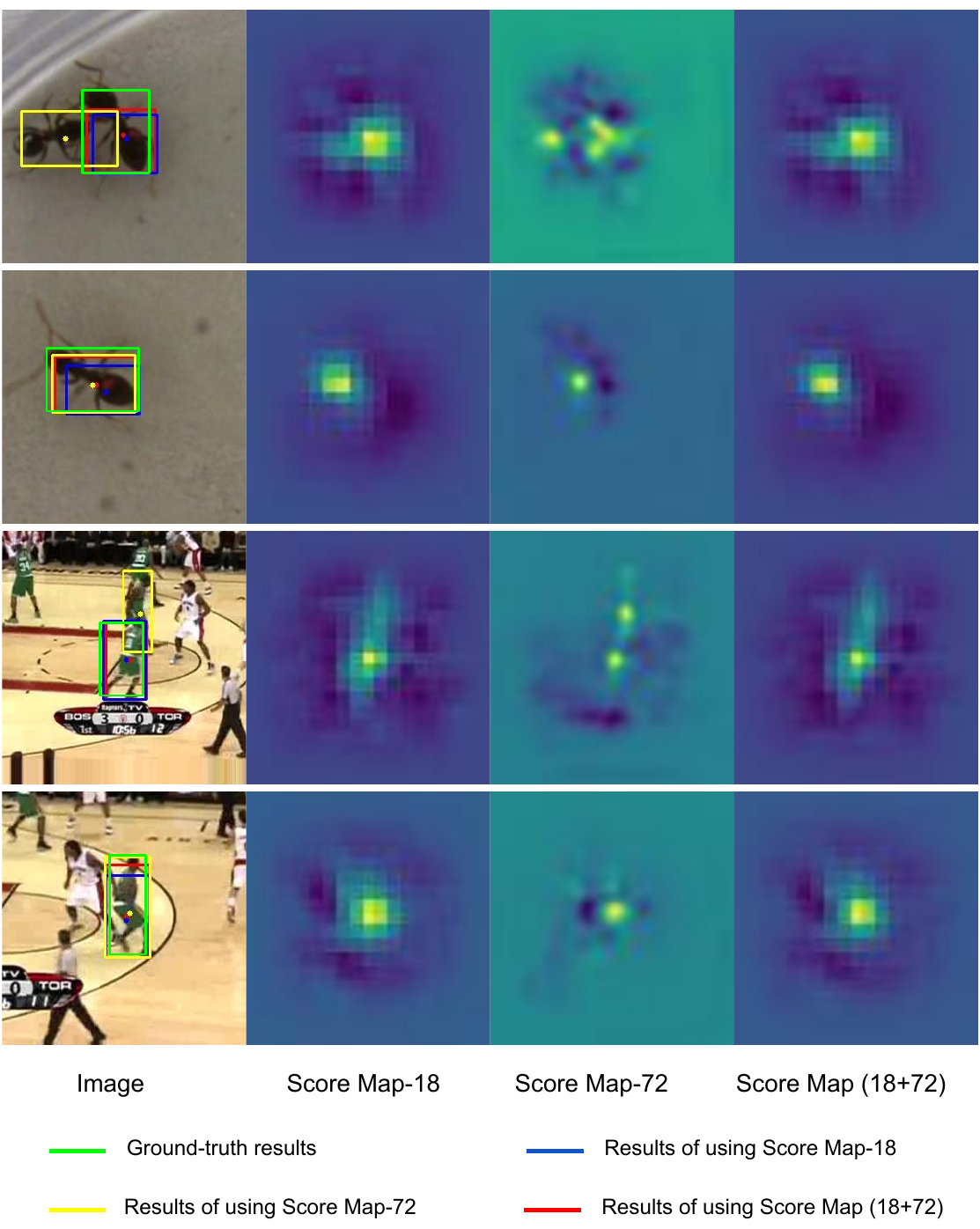}
\caption{Visualization of score maps and results generated by using different layers of classification branch in our FCOT on VOT2018 dataset~\cite{vot2018}. In the images of the first column, green box denotes the ground-truth bounding box of the target, red box and dot represent for the bounding box and the center of the target generated by using multi-scale classification strategy, yellow for solely using Score Map-72 and blue for solely using Score Map-72.
Best viewed with zooming in.}
\label{fig:sup2}
\end{figure}

\paragraph{UAV123}
UAV123~\cite{uav123} is a large dataset captured from low-altitude UAVs with sequences from an aerial viewpoint, a subset of which is meant for long-term aerial tracking. Thus compared with other benchmarks, the targets tend to be farther from the camera in UAV123. This dataset contains a total of 123 video sequences and more than 110K frames. We compare our FCOT with previous approaches on this dataset as in Table~\ref{table:uav}. FCOT outperforms the previous best approaches reaching 65.6\% in AUC score and 87.3\% in precision score, respectively. 

\paragraph{NFS}

NFS dataset~\cite{nfs} contains a total of 380K frames in 100 videos captured with high frame rate cameras from real world scenarios. We evaluate our FCOT on the 30 FPS version of this dataset and compare with the recent approaches. The results are shown on Table~\ref{table:nfs}. Specifically, FCOT obtains the highest AUC score of 62.7\% and precision score of 75.4\%.

\paragraph{OTB-100}

We evaluate our FCOT on the 100 videos of OTB-100~\cite{otb} dataset, with variable challenging aspects including occlusion, deformation, out-of-plane rotation, low resolution, etc. Fig~\ref{fig:otb} reports our overall performance of FCOT on OTB-100. We achieve 69.9\% and 92.2\% in success score and precision score, which outperforms DiMP with relative gains of 0.9\% and 1.9\% in the two metrics and surpass other anchor-free trackers including OCEAN, SiamFC++, etc.

\subsection{Qualitative Analysis}

\paragraph{Visualization results on OTB100}
We provide visualization examples generated by FCOT and DiMP on OTB100 dataset~\cite{otb} in Figure~\ref{fig:sup1}. These sequences suffer from the limitations including occlusion, scale variation, deformation, motion blur and so on. It can be seen that our tracker performs well on these sequences. Particularly, the bounding boxes are more precise than DiMP~\cite{dimp} once the objects have been roughly located by the classification branch.

\paragraph{Visualization of Classification heat-maps at Different Layers}
In this section, we compare the tracking results and score maps among applying different-scale classification strategies to verify the effectiveness of multi-scale classification strategy as in Figure~\ref{fig:sup2}. We can see from the first and the third row that the results of just using Score Map-72 are deviated from the ground truth while trackers that using just Score Map-18 and using both of them can discriminate the positive object from the similar ones. It demonstrates that Score Map-18 is helpful for the robustness of the tracker. While from the second and the last row, we can derive that the predicted bounding boxes and centers of using Score Map-72 are more precise than only using Score Map-18. In consequence, multi-scale classification strategy is helpful for both the robustness and accuracy.

\section{Conclusions and Future Work}
In this paper, we have presented a fully convolutional online tracker (FCOT), by unifying the components of feature extraction, classification head, and regression head into a encoder-decoder architecture. Our key contribution is to introduce an online regression model generator (RMG) based on the carefully designed anchor-free box regression branch, which enables our FCOT to be more effective in handling target deformation during tracking procedure.
In addition, to deal with the confusion of similar objects, we devise a simple yet effective multi-scale classification branch. Extensive experiments on several benchmarks demonstrate the high precision of our proposed online anchor-free regression branch and multi-scale classification. Our FCOT outperforms the state-of-the-art trackers on most benchmarks at a high speed of over 40 FPS.

Relatively large training cost due to the existence of online module and demands for intricate hyper-parameters tuning is the main drawback of our method. To overcome that, we consider to improve our method from two aspects. First, we could search for a more efficient and simple backbone to compress the architecture. Second, we plan to explore a more effective online module for classification and regression with fewer parameters.


\ifCLASSOPTIONcaptionsoff
  \newpage
\fi



%

\bibliographystyle{splncs04}
\bibliography{egbib}

\begin{thebibliography}{10}
\providecommand{\url}[1]{\texttt{#1}}
\providecommand{\urlprefix}{URL }
\providecommand{\doi}[1]{https://doi.org/#1}

\bibitem{staple}
Bertinetto, L., Valmadre, J., Golodetz, S., Miksik, O., Torr, P.H.S.: Staple:
  Complementary learners for real-time tracking. In: Proceedings of the IEEE
  Conference on Computer Vision and Pattern Recognition (CVPR) (June 2016)

\bibitem{siamfc}
Bertinetto, L., Valmadre, J., Henriques, J.F., Vedaldi, A., Torr, P.H.S.:
  Fully-convolutional siamese networks for object tracking. In: Hua, G.,
  J{\'{e}}gou, H. (eds.) {ECCV} Workshops (2016)

\bibitem{dimp}
Bhat, G., Danelljan, M., Gool, L.V., Timofte, R.: Learning discriminative model
  prediction for tracking. CoRR  (2019)

\bibitem{updt}
Bhat, G., Johnander, J., Danelljan, M., Khan, F.S., Felsberg, M.: Unveiling the
  power of deep tracking. In: Ferrari, V., Hebert, M., Sminchisescu, C., Weiss,
  Y. (eds.) {ECCV} (2018)

\bibitem{tt}
Chen, X., Yan, B., Zhu, J., Wang, D., Yang, X., Lu, H.: Transformer tracking.
  In: Proceedings of the IEEE/CVF Conference on Computer Vision and Pattern
  Recognition (CVPR). pp. 8126--8135 (June 2021)

\bibitem{siamban}
Chen, Z., Zhong, B., Li, G., Zhang, S., Ji, R.: Siamese box adaptive network
  for visual tracking. In: {CVPR} (June 2020)

\bibitem{treg}
Cui, Y., Jiang, C., Wang, L., Wu, G.: Target transformed regression for
  accurate tracking (2021)

\bibitem{deformable_conv}
Dai, J., Qi, H., Xiong, Y., Li, Y., Zhang, G., Hu, H., Wei, Y.: Deformable
  convolutional networks. In: {ICCV} (2017)

\bibitem{eco}
Danelljan, M., Bhat, G., Khan, F.S., Felsberg, M.: {ECO:} efficient convolution
  operators for tracking. In: {CVPR} (2017)

\bibitem{atom}
Danelljan, M., Bhat, G., Khan, F.S., Felsberg, M.: {ATOM:} accurate tracking by
  overlap maximization. In: {CVPR} (2019)

\bibitem{srdcf}
Danelljan, M., Hager, G., Shahbaz~Khan, F., Felsberg, M.: Learning spatially
  regularized correlation filters for visual tracking. In: Proceedings of the
  IEEE International Conference on Computer Vision (ICCV) (December 2015)

\bibitem{dsst}
Danelljan, M., Häger, G., Shahbaz~Khan, F., Felsberg, M.: Accurate scale
  estimation for robust visual tracking. In: Proceedings of the British Machine
  Vision Conference. BMVA Press (2014)

\bibitem{ccot}
Danelljan, M., Robinson, A., Khan, F.S., Felsberg, M.: Beyond correlation
  filters: Learning continuous convolution operators for visual tracking. In:
  Leibe, B., Matas, J., Sebe, N., Welling, M. (eds.) {ECCV} (2016)

\bibitem{samf}
Danelljan, M., Shahbaz~Khan, F., Felsberg, M., van~de Weijer, J.: Adaptive
  color attributes for real-time visual tracking. In: Proceedings of the IEEE
  Conference on Computer Vision and Pattern Recognition (CVPR) (June 2014)

\bibitem{prdimp}
Danelljan, M., Van~Gool, L., Timofte, R.: Probabilistic regression for visual
  tracking. In: {CVPR} (2020)

\bibitem{CGACD}
Du, F., Liu, P., Zhao, W., Tang, X.: Correlation-guided attention for corner
  detection based visual tracking. In: {CVPR} (June 2020)

\bibitem{centernet}
Duan, K., Bai, S., Xie, L., Qi, H., Huang, Q., Tian, Q.: Centernet: Keypoint
  triplets for object detection. In: Proceedings of the IEEE/CVF International
  Conference on Computer Vision (ICCV) (October 2019)

\bibitem{lasot}
Fan, H., Lin, L., Yang, F., Chu, P., Deng, G., Yu, S., Bai, H., Xu, Y., Liao,
  C., Ling, H.: Lasot: {A} high-quality benchmark for large-scale single object
  tracking. In: {CVPR} (2019)

\bibitem{CRPN}
Fan, H., Ling, H.: Siamese cascaded region proposal networks for real-time
  visual tracking. In: {CVPR} (2019)

\bibitem{Siamese_Cascaded_Region_Proposal_Networks_for_Real_Time_Visual_Tracking}
Fan, H., Ling, H.: Siamese cascaded region proposal networks for real-time
  visual tracking. In: {CVPR} (2019)

\bibitem{stmtrack}
Fu, Z., Liu, Q., Fu, Z., Wang, Y.: Stmtrack: Template-free visual tracking with
  space-time memory networks. In: Proceedings of the IEEE/CVF Conference on
  Computer Vision and Pattern Recognition (CVPR). pp. 13774--13783 (June 2021)

\bibitem{nfs}
Galoogahi, H.K., Fagg, A., Huang, C., Ramanan, D., Lucey, S.: Need for speed: A
  benchmark for higher frame rate object tracking. arXiv preprint
  arXiv:1703.05884  (2017)

\bibitem{satin}
Gao, P., Yuan, R., Wang, F., Xiao, L., Fujita, H., Zhang, Y.: Siamese
  attentional keypoint network for high performance visual tracking.
  Knowledge-Based Systems  \textbf{193},  105448 (Apr 2020).
  \doi{10.1016/j.knosys.2019.105448},
  \url{http://dx.doi.org/10.1016/j.knosys.2019.105448}

\bibitem{siamcar}
Guo, D., Wang, J., Cui, Y., Wang, Z., Chen, S.: Siamcar: Siamese fully
  convolutional classification and regression for visual tracking. In: {CVPR}
  (June 2020)

\bibitem{Learning_dynamic}
Guo, Q., Feng, W., Zhou, C., Huang, R., Wan, L., Wang, S.: Learning dynamic
  siamese network for visual object tracking. In: {ICCV} (2017)

\bibitem{resnet50}
He, K., Zhang, X., Ren, S., Sun, J.: Deep residual learning for image
  recognition. In: {CVPR} (2016)

\bibitem{kcf}
Henriques, J.F., Caseiro, R., Martins, P., Batista, J.: High-speed tracking
  with kernelized correlation filters. {IEEE} Trans. Pattern Anal. Mach.
  Intell.  \textbf{37}(3),  583--596 (2015)

\bibitem{6870486}
Henriques, J.F., Caseiro, R., Martins, P., Batista, J.: High-speed tracking
  with kernelized correlation filters. IEEE Transactions on Pattern Analysis
  and Machine Intelligence  \textbf{37}(3),  583--596 (2015).
  \doi{10.1109/TPAMI.2014.2345390}

\bibitem{got10k}
Huang, L., Zhao, X., Huang, K.: Got-10k: {A} large high-diversity benchmark for
  generic object tracking in the wild. CoRR  (2018)

\bibitem{iounet}
Jiang, B., Luo, R., Mao, J., Xiao, T., Jiang, Y.: Acquisition of localization
  confidence for accurate object detection. In: Ferrari, V., Hebert, M.,
  Sminchisescu, C., Weiss, Y. (eds.) {ECCV} (2018)

\bibitem{Real_Time_Object_Tracking}
Jung, I., You, K., Noh, H., Cho, M., Han, B.: Real-time object tracking and
  one-shot channel pruning via meta-learning: Efficient model adaptation. CoRR
  (2019)

\bibitem{Galoogahi_2017_ICCV}
Kiani~Galoogahi, H., Fagg, A., Lucey, S.: Learning background-aware correlation
  filters for visual tracking. In: Proceedings of the IEEE International
  Conference on Computer Vision (ICCV) (Oct 2017)

\bibitem{bacf}
Kiani~Galoogahi, H., Fagg, A., Lucey, S.: Learning background-aware correlation
  filters for visual tracking. In: Proceedings of the IEEE International
  Conference on Computer Vision (ICCV) (Oct 2017)

\bibitem{adam}
Kingma, D.P., Ba, J.: Adam: {A} method for stochastic optimization. In: Bengio,
  Y., LeCun, Y. (eds.) {ICLR} (2015)

\bibitem{vot2018}
Kristan, M., Leonardis, A., Matas, J., Felsberg, M., Pflugfelder, R.P., Zajc,
  L.C., et~al: The sixth visual object tracking {VOT2018} challenge results.
  In: {ECCV} Workshops (2018)

\bibitem{alexnet}
Krizhevsky, A., Sutskever, I., Hinton, G.E.: Imagenet classification with deep
  convolutional neural networks. Commun. {ACM}  \textbf{60}(6),  84--90 (2017)

\bibitem{siamrpnPlus}
Li, B., Wu, W., Wang, Q., Zhang, F., Xing, J., Yan, J.: Siamrpn++: Evolution of
  siamese visual tracking with very deep networks. In: {CVPR} (2019)

\bibitem{siamrpn}
Li, B., Yan, J., Wu, W., Zhu, Z., Hu, X.: High performance visual tracking with
  siamese region proposal network. In: {CVPR} (2018)

\bibitem{strcf}
Li, F., Tian, C., Zuo, W., Zhang, L., Yang, M.H.: Learning spatial-temporal
  regularized correlation filters for visual tracking. In: Proceedings of the
  IEEE Conference on Computer Vision and Pattern Recognition (CVPR) (June 2018)

\bibitem{introduction1}
Liu, L., Xing, J., Ai, H., Ruan, X.: Hand posture recognition using finger
  geometric feature. In: {ICPR} (2012)

\bibitem{dcf_}
Lukezic, A., Vojir, T., Zajc, L.C., Matas, J., Kristan, M.: Discriminative
  correlation filter with channel and spatial reliability. In: {CVPR} (2017)

\bibitem{hcf}
Ma, C., Huang, J.B., Yang, X., Yang, M.H.: Hierarchical convolutional features
  for visual tracking. In: Proceedings of the IEEE International Conference on
  Computer Vision (ICCV) (December 2015)

\bibitem{goturn}
Ma, C., Huang, J.B., Yang, X., Yang, M.H.: Hierarchical convolutional features
  for visual tracking. In: 2015 IEEE International Conference on Computer
  Vision (ICCV). pp. 3074--3082 (2015). \doi{10.1109/ICCV.2015.352}

\bibitem{pytracking}
Martin~Danelljan, G.B.: pytracking.
  \url{https://github.com/visionml/pytracking}

\bibitem{uav123}
Mueller, M., Smith, N., Ghanem, B.: A benchmark and simulator for {UAV}
  tracking. In: Leibe, B., Matas, J., Sebe, N., Welling, M. (eds.) {ECCV}
  (2016)

\bibitem{trackingnet}
M{\"{u}}ller, M., Bibi, A., Giancola, S., Al{-}Subaihi, S., Ghanem, B.:
  Trackingnet: {A} large-scale dataset and benchmark for object tracking in the
  wild. In: Ferrari, V., Hebert, M., Sminchisescu, C., Weiss, Y. (eds.) {ECCV}
  (2018)

\bibitem{mdnet}
Nam, H., Han, B.: Learning multi-domain convolutional neural networks for
  visual tracking. In: {CVPR} (2016)

\bibitem{Nam_2016_CVPR}
Nam, H., Han, B.: Learning multi-domain convolutional neural networks for
  visual tracking. In: Proceedings of the IEEE Conference on Computer Vision
  and Pattern Recognition (CVPR) (June 2016)

\bibitem{numerical_optimization}
Nocedal, J., Wright, S.J.: Numerical Optimization. Springer (1999)

\bibitem{meta-tracker}
Park, E., Berg, A.C.: Meta-tracker: Fast and robust online adaptation for
  visual object trackers. In: Ferrari, V., Hebert, M., Sminchisescu, C., Weiss,
  Y. (eds.) {ECCV} (2018)

\bibitem{Park_2018_ECCV}
Park, E., Berg, A.C.: Meta-tracker: Fast and robust online adaptation for
  visual object trackers. In: Proceedings of the European Conference on
  Computer Vision (ECCV) (September 2018)

\bibitem{hdt}
Qi, Y., Zhang, S., Qin, L., Yao, H., Huang, Q., Lim, J., Yang, M.H.: Hedged
  deep tracking. In: Proceedings of the IEEE Conference on Computer Vision and
  Pattern Recognition (CVPR) (June 2016)

\bibitem{faster_rcnn}
Ren, S., He, K., Girshick, R.B., Sun, J.: Faster {R-CNN:} towards real-time
  object detection with region proposal networks. In: Advances in Neural
  Information Processing Systems 28: Annual Conference on Neural Information
  Processing Systems 2015, December 7-12, 2015, Montreal, Quebec, Canada (2015)

\bibitem{fcos}
Tian, Z., Shen, C., Chen, H., He, T.: {FCOS:} fully convolutional one-stage
  object detection. CoRR  (2019)

\bibitem{Valmadre_2017_CVPR}
Valmadre, J., Bertinetto, L., Henriques, J., Vedaldi, A., Torr, P.H.S.:
  End-to-end representation learning for correlation filter based tracking. In:
  Proceedings of the IEEE Conference on Computer Vision and Pattern Recognition
  (CVPR) (July 2017)

\bibitem{siamrcnn}
Voigtlaender, P., Luiten, J., Torr, P.H., Leibe, B.: Siam r-cnn: Visual
  tracking by re-detection. In: Proceedings of the IEEE/CVF Conference on
  Computer Vision and Pattern Recognition (CVPR) (June 2020)

\bibitem{spm_tracker}
Wang, G., Luo, C., Xiong, Z., Zeng, W.: Spm-tracker: Series-parallel matching
  for real-time visual object tracking. In: Proceedings of the IEEE/CVF
  Conference on Computer Vision and Pattern Recognition (CVPR) (June 2019)

\bibitem{tmt}
Wang, N., Zhou, W., Wang, J., Li, H.: Transformer meets tracker: Exploiting
  temporal context for robust visual tracking. In: Proceedings of the IEEE/CVF
  Conference on Computer Vision and Pattern Recognition (CVPR). pp. 1571--1580
  (June 2021)

\bibitem{otb}
Wu, Y., Lim, J., Yang, M.: Object tracking benchmark. {IEEE} Trans. Pattern
  Anal. Mach. Intell.  \textbf{37}(9),  1834--1848 (2015)

\bibitem{introduction2}
Xing, J., Ai, H., Lao, S.: Multiple human tracking based on multi-view
  upper-body detection and discriminative learning. In: {ICPR} (2010)

\bibitem{gfs_dcf}
Xu, T., Feng, Z.H., Wu, X.J., Kittler, J.: Joint group feature selection and
  discriminative filter learning for robust visual object tracking. In:
  Proceedings of the IEEE/CVF International Conference on Computer Vision
  (ICCV) (October 2019)

\bibitem{siamfc++}
Xu, Y., Wang, Z., Li, Z., Ye, Y., Yu, G.: Siamfc++: Towards robust and accurate
  visual tracking with target estimation guidelines. CoRR  (2019)

\bibitem{Yao_2018_ECCV}
Yao, Y., Wu, X., Zhang, L., Shan, S., Zuo, W.: Joint representation and
  truncated inference learning for correlation filter based tracking. In:
  Proceedings of the European Conference on Computer Vision (ECCV) (September
  2018)

\bibitem{meem}
Zhang, J., Ma, S., Sclaroff, S.: {MEEM:} robust tracking via multiple experts
  using entropy minimization. In: Proc. of the European Conference on Computer
  Vision (ECCV) (2014)

\bibitem{StructSiam}
Zhang, Y., Wang, L., Qi, J., Wang, D., Feng, M., Lu, H.: Structured siamese
  network for real-time visual tracking. In: Ferrari, V., Hebert, M.,
  Sminchisescu, C., Weiss, Y. (eds.) {ECCV} (2018)

\bibitem{ocean}
Zhang, Z., Peng, H., Fu, J., Li, B., Hu, W.: Ocean: Object-aware anchor-free
  tracking. In: {ECCV} (2020)

\bibitem{deformable_conv_v2}
Zhu, X., Hu, H., Lin, S., Dai, J.: Deformable convnets {V2:} more deformable,
  better results. In: {CVPR} (2019)

\bibitem{dasiamrpn}
Zhu, Z., Wang, Q., Li, B., Wu, W., Yan, J., Hu, W.: Distractor-aware siamese
  networks for visual object tracking. In: Ferrari, V., Hebert, M.,
  Sminchisescu, C., Weiss, Y. (eds.) {ECCV} (2018)

\end{thebibliography}
\end{document}